\documentclass[10pt,twocolumn,letterpaper]{article}

\usepackage{iccv}
\usepackage{times}
\usepackage{epsfig}
\usepackage{graphicx}
\usepackage{amsmath}
\usepackage{amssymb}
\usepackage{bm}
\usepackage{multirow}
\usepackage{makecell}
\usepackage{bbding}

\usepackage[breaklinks=true,bookmarks=false]{hyperref}

\iccvfinalcopy 

\def\iccvPaperID{****} 

\ificcvfinal\fi

\begin{document}

\title{Foreground-Action Consistency Network for Weakly Supervised Temporal Action Localization}

\author{Linjiang Huang\textsuperscript{\rm 1,2} \quad
        Liang Wang\textsuperscript{\rm 3} \quad
        Hongsheng Li\textsuperscript{\rm 1,2} \thanks{Corresponding author.}\\
\textsuperscript{\rm 1}Multimedia Laboratory, The Chinese University of Hong Kong \\
\textsuperscript{\rm 2}Centre for Perceptual and Interactive Intelligence, Hong Kong \\
\textsuperscript{\rm 3}Institute of Automation, Chinese Academy of Sciences \\
{\tt\small ljhuang524@gmail.com, wangliang@nlpr.ia.ac.cn, hsli@ee.cuhk.edu.hk}
}

\maketitle
\ificcvfinal\thispagestyle{empty}\fi

\begin{abstract}
As a challenging task of high-level video understanding, weakly supervised temporal action localization has been attracting increasing attention. With only video annotations, most existing methods seek to handle this task with a localization-by-classification framework, which generally adopts a selector to select snippets of high probabilities of actions or namely the foreground. Nevertheless, the existing foreground selection strategies have a major limitation of only considering the unilateral relation from foreground to actions, which cannot guarantee the foreground-action consistency. In this paper, we present a framework named FAC-Net based on the I3D backbone, on which three branches are appended, named class-wise foreground classification branch, class-agnostic attention branch and multiple instance learning branch.
First, our class-wise foreground classification branch regularizes the relation between actions and foreground to maximize the foreground-background separation.
Besides, the class-agnostic attention branch and multiple instance learning branch are adopted to regularize the foreground-action consistency and help to learn a meaningful foreground classifier.
Within each branch, we introduce a hybrid attention mechanism, which calculates multiple attention scores for each snippet, to focus on both discriminative and less-discriminative snippets to capture the full action boundaries. Experimental results on THUMOS14 and ActivityNet1.3 demonstrate the state-of-the-art performance of our method. Our code is available at \url{https://github.com/LeonHLJ/FAC-Net}.
\end{abstract}

\section{Introduction}
Temporal action localization in videos has been widely used in various fields \cite{sun2015temporal,sultani2018real}. This task aims to localize action instances in untrimmed videos along the temporal dimension. Most existing methods \cite{yeung2016end,shou2016temporal,xu2017r,zhao2017temporal,chao2018rethinking,lin2018bsn,long2019gaussian} are trained in a fully supervised manner. However, such a requirement of frame-level annotations does not suit real-world applications since densely annotating large-scale videos is expensive and time-consuming.
To address this difficulty, weakly supervised methods \cite{laptev2008learning,bojanowski2014weakly,wang2017untrimmednets} have been developed with only video-level labels, which are much easier to annotate. Among diverse weak supervisions, video category labels are the most commonly used \cite{wang2017untrimmednets,nguyen2018weakly,paul2018w}.

\begin{figure}[!t]
\centering
\includegraphics[width=0.95\columnwidth]{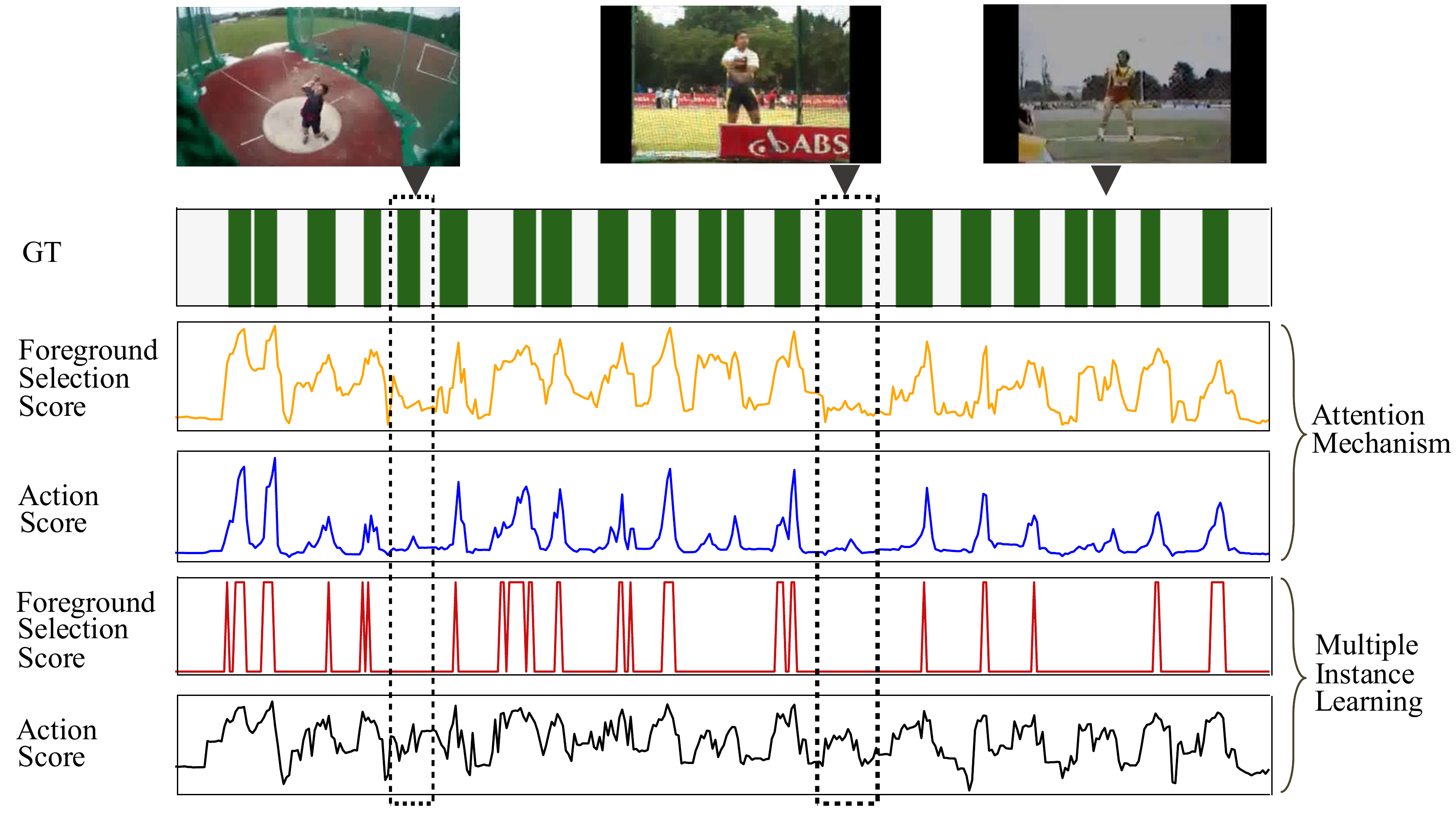}
\caption{An example of the action of ``\emph{HammerThrow}''. The barcode is the ground-truth (GT). The following line charts are foreground selection scores and frame-wise classification scores. We show the results of STPN \cite{paul2018w} (attention mechanism) and W-TALC \cite{nguyen2018weakly} (multiple instance learning). It is obvious that these methods cannot guarantee foreground-action consistency.}\label{fig:motivation}
\vspace{-6mm}
\end{figure}

Because of the absence of frame-wise annotations, existing works mainly embrace a localization-by-classification pipeline \cite{wang2017untrimmednets,yang2020equivalent}, in which an important component is the selector to select snippets with high probabilities of actions, or namely the {\it foreground}. Existing mechanisms for foreground selection can be categorized into two main strategies, \ie, attention mechanism \cite{nguyen2018weakly,liu2019completeness} and multiple instance learning (MIL) \cite{paul2018w}. However, both strategies have their intrinsic drawbacks.
As shown in Fig. \ref{fig:motivation}, the attention mechanism (either class-agnostic attention or class-wise attention) usually suffers from the discrepancy \cite{singh2017hide,zhong2018step,liu2019completeness} between classification and detection, \ie, the attention scores being concentrated around most discriminative action snippets or wrongly focus on background snippets.
On the other hand, the multiple instance learning should rely on a temporal top-$k$ pooling operation, but there is no guarantee that all the top-$k$ snippets would be foreground, since the number $k$ is generally defined by humans. In summary, existing approaches lack the ability to maintain consistency between foreground and actions, that is, the foreground and actions should be mutually inclusive.

In this work, we propose to tackle the action localization problem by explicitly modeling and regularizing the foreground-action consistency. Given the insight that existing foreground selection strategies only consider the unilateral relation from foreground to actions, we propose a framework to further take bilateral relations into consideration. Based on a common video backbone, our method appends three branches on top of it. The first branch, named as class-wise foreground classification branch (CW branch, Sec.\ref{sec:cw_branch}), seeks to model the action-to-foreground relation. Meanwhile, it acts similarly to the noise contrastive estimation (NCE) \cite{gutmann2010noise,oord2018representation}, which actually maximizes a lower bound on mutual information (MI) between the foreground feature and the feature of ground truth action, leading to better foreground-background separation. The second branch (CA branch, Sec.\ref{sec:ca_branch}) introduces a class-agnostic attention mechanism that to model the reverse foreground-to-action relation for complementing the first branch, so as to build the foreground-action consistency. Moreover, it enables to learn a semantically meaningful foreground feature. The third branch (MIL branch, Sec.\ref{sec:mil_branch}) is an MIL-like pipeline to further improve video classification and facilitate the learning of class-wise attention in the CW branch.

Within each branch, we adopt a hybrid attention mechanism to ease the attention learning and promote precise foreground prediction. In addition to focusing on key frames in the video, the hybrid attention mechanism can learn to accommodate less-discriminative snippets, which benefits capturing accurate action boundaries.
To evaluate the effectiveness of our method, we perform experiments on two benchmarks, THUMOS14 \cite{THUMOS14} and ActivityNet1.3 \cite{caba2015activitynet}. Experimental results on the two benchmarks demonstrate the superior performance over state-of-the-art approaches.

Our main contributions are three-fold.
(a) We introduce a class-wise foreground classification pipeline to improve the robustness of foreground prediction. This pipeline models and regularizes the foreground-action consistency that is mostly ignored by existing methods.
(b) We propose a hybrid attention mechanism to improve the attention learning and help to capture accurate action boundaries.
(c) The proposed class-wise foreground classification pipeline can play a complementary role over existing methods to consistently improve the action localization performance.

\section{Related Work}
\vspace{-1mm}
\paragraph{Fully Supervised Temporal Action Localization.}
Different from action recognition \cite{karpathy2014large,simonyan2014two,tran2015learning,carreira2017quo}, temporal action localization aims to localize the start and end points of action instances, meanwhile recognizing the action category of each action instance.
We group fully-supervised methods into two categories. The methods in the first category employ a multi-stage pipeline including proposal generation, classification and proposal refinement.
These methods mainly focus on improving the quality of proposals \cite{dai2017temporal,lin2018bsn} and learning robust and accurate classifiers \cite{shou2016temporal,zhao2017temporal}. In the second category, methods aim to generate action labels at the frame-level granularity \cite{shou2017cdc,lea2017temporal,yuan2017temporal}, which need an additional merging step to obtain the final temporal boundaries. The rest of the methods mainly rely on end-to-end architectures. Even though these methods have achieved promising performance, they severely rely on frame-wise annotations.

\vspace{-4mm}
\paragraph{Weakly Supervised Temporal Action Localization.}
Recently, many attempts have been made to solve temporal action localization with weak labels. UntrimmedNet \cite{wang2017untrimmednets} proposed to tackle this problem by selecting
relevant segments with attention mechanism or multiple instance learning, which is followed by most of the subsequent methods.
\begin{figure*}[t]
\centering
\includegraphics[width=0.93\textwidth]{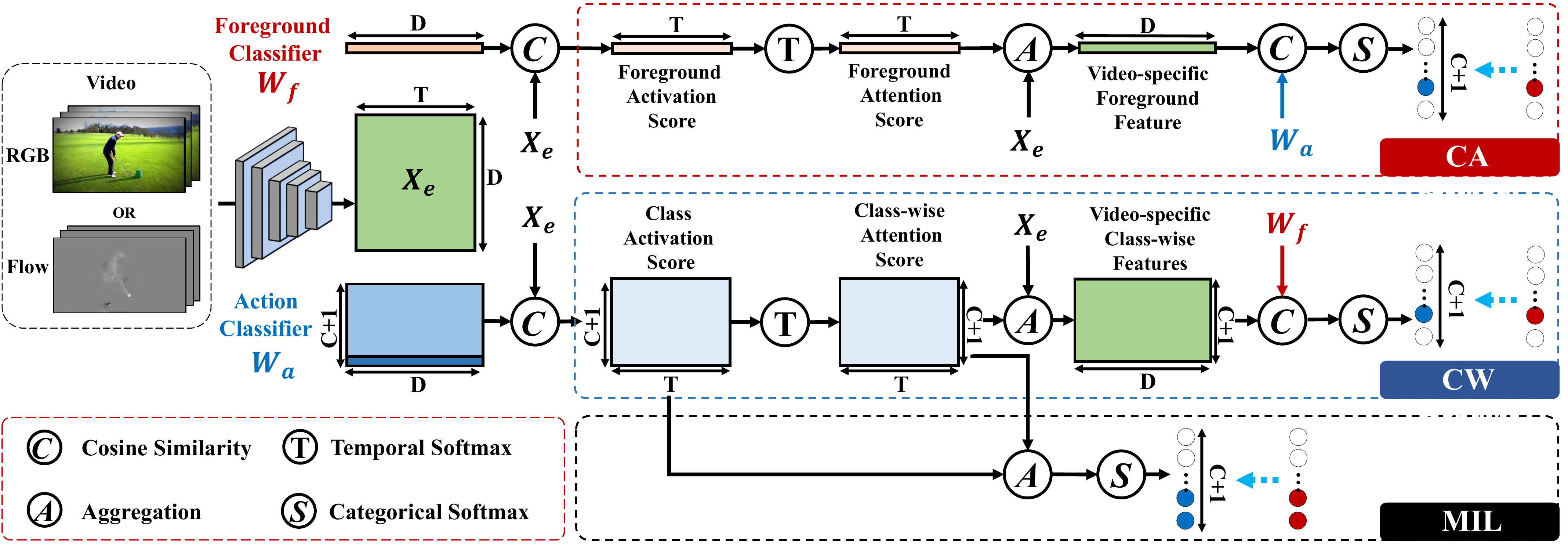}
\caption{The overview of our method. We have three main branches. The class-wise foreground classification branch (CW branch) seeks to build the relation from actions to foreground, while the class-agnostic attention pipeline (CA branch) complements the reverse relation from foreground to actions for the first branch, so as to build the foreground-action consistency. The multiple instance learning branch (MIL branch) is a MIL-like pipeline to provide a different perspective for video classification and facilitate the learning of class-wise attention. For more details about the hybrid attention strategy, please refer to Fig. \ref{fig:hybrid_attention}.}
\label{fig:framework}
\vspace{-3mm}
\end{figure*}

Attention-based methods aim to select snippets of high action probabilities by using attention mechanism. For instance, STPN \cite{nguyen2018weakly} introduced a sparse regularization on the attention sequence to capture key frames of a video. 3C-Net \cite{narayan20193c} proposed to learn class-wise attention to obtain class-wise features for calculating a center loss. Some methods \cite{nguyen2019weakly,moniruzzaman2020action,huang2021modeling} leveraged the complementary nature of foreground and background to generate both foreground and background attention sequences to explicitly model background. HAM-Net \cite{islam2021hybrid} proposed a hybrid attention mechanism that includes
temporal soft, semi-soft and hard attentions to capture full action instances. Note that, our method also utilize a hybrid attention mechanism, but totally differs from HAM-Net. First, we utilize different temperature values to generate multiple foreground attention sequences, while HAM-Net generate soft, semi-soft and hard attention sequences mainly by thresholding. Second, we only generate foreground attention sequences, while the hard attention sequence of HAM-Net also contain background snippets.

MIL-based methods \cite{paul2018w,lee2019background,moniruzzaman2020action} can be regarded as a hard selection mechanism based on the principle of multiple instance learning. In contrast to attention-based methods that automatically learn attention weights, MIL-based methods mainly rely on a top-$k$ selection operation to select positive instances in the bag (video). However, as stated above, neither attention-based methods nor MIL-based methods can maintain foreground-action consistency.

There are some methods have noticed the importance of foreground-action consistency. For example, RefineLoc \cite{pardo2021refineloc} generated snippet-level hard pseudo labels by expanding previous detection results, TSCN \cite{zhai2020two} generated pseudo ground truth from the foreground attention sequence, and EM-MIL \cite{luo2020weakly} put the pseudo-label generation into an expectation-maximization framework. There are also some methods \cite{nguyen2019weakly,moniruzzaman2020action,islam2021hybrid} attempted to use class activations as a \emph{top-down} supervision to guide foreground attention generation. The differences between our method and these methods are that we require no additional supervision to enforce foreground-action consistency, and in theory our method can achieve better foreground-background separation.

\section{Proposed Method} \label{sec:proposed_method}
In this section, we elaborate on the proposed method. The overview of the proposed method is shown in Fig.  \ref{fig:framework}.

\vspace{-4mm}
\paragraph{Problem definition.}
Let \(V=\{v_t\}_{t=1}^{L}\) be a video of temporal length \(L\). Assume that we have a set of \(N\) training videos \(\{V_i\}_{i=1}^{N}\) which are annotated with their action categories \(\{\bm{y}_i\}_{i=1}^{N}\), where \(\bm{y}_i\) is a binary vector indicating the presence/absence of each action. During inference, for a video, we predict a set of action instances \(\{(c, q, t_s, t_e)\}\), where \(c\) denotes the predicted action class, \(q\) is the confidence score, \(t_s\) and \(t_e\) represent the start time and end time of the instance.

\vspace{-4mm}
\paragraph{Relation definition.} (1) \emph{Foreground-to-action relation}: the unilateral relation from foreground to actions, \ie, the foreground must be some kind of actions. (2) \emph{Action-to-foreground relation}: the unilateral relation from actions to foreground, \ie, actions must be the foreground.

\vspace{-4mm}
\paragraph{Overview.} There are four modules in our method. Based on the I3D backbone, we utilize a feature embedding module (Sec.\ref{sec:feature_embedding}) to extract task-oriented features. Then, three branches are appended on the top of it. The first branch, named as class-wise foreground classification branch (CW branch, Sec.\ref{sec:cw_branch}), seeks to model the action-to-foreground relation. The second branch (CA branch, Sec.\ref{sec:ca_branch}) introduces a class-agnostic attention mechanism to model the reverse foreground-to-action relation for complementing the first branch, so as to build the foreground-action consistency. The third branch (MIL branch, Sec.\ref{sec:mil_branch}) is an MIL-like pipeline to further improve video classification and facilitate the learning of class-wise attention in the CW branch.

\subsection{Feature Embedding Module} \label{sec:feature_embedding}
In order to extract task-specific features for temporal action localization, we first utilize a feature embedding module that comprises two parts. The first part is a pre-trained network, \ie, I3D \cite{carreira2017quo}. Given a video, we first extract RGB features and optical-flow features respectively by the fixed backbone network. After feature encoding, we employ a two-layer temporal convolutional network \cite{paul2018w,liu2019completeness} to learn task-oriented features \(\bm{X}_{e} \in \mathbb{R}^{T \times D}\), where \(T\) denotes the number of snippets, and \(D\) is the dimension.

\subsection{Class-Wise Foreground Classification Branch} \label{sec:cw_branch}
As mentioned above, foreground and actions should be consistent and mutually inclusive. Nevertheless, most existing methods only consider the foreground-to-action relation, in other words, they only take advantage of the prior that the foreground must be actions. A possible result is that the obtained foreground scores only focus on discriminative action snippets, which also conforms to this unilateral relation. Intuitively, a rational relation between foreground and actions should be bilateral, it is essential to further take the action-to-foreground relation into consideration.
Inspired by the class-agnostic attention pipeline \cite{nguyen2018weakly,liu2019completeness,lee2019background} that exploits the unilateral relation from foreground to actions, we propose a symmetric pipeline named class-wise foreground classification pipeline as a branch in our method.

We first randomly initialize an action classifier \(\bm{W}_{a} \in \mathbb{R}^{(C+1) \times D}\) and a foreground classifier \(\bm{W}_{f} \in \mathbb{R}^{D}\), where \(C\) denotes the number of action categories, and the ($C+1$)-th class corresponds to the background.
Given the embedding \(\bm{X}_{e}\), we calculate the cosine similarities between \(\bm{X}_{e}\) and \(\bm{W}_{a}\) to obtain the frame-wise class activation scores \(\bm{S}_a \in \mathbb{R}^{T \times (C+1)} \) as
\begin{equation} \label{eq:class_similaritiy}
\bm{S}_a(t, j) = \delta \cdot \cos(\bm{X}_e(t), \bm{W}_{a}(j)),
\end{equation}
where \(\bm{X}_e(t)\) denotes the embedding of the \(t^{th}\) snippet, and \(\delta\) is a scalar to control the scale of the value.

To build the action-to-foreground relation, we follow the line that action snippets are also foreground snippets and calculate the class-wise attention scores \(\bm{A}_a \in \mathbb{R}^{T \times (C+1)} \), which are used to aggregate the embedding \(\bm{X}_e\) into the video-specific class-wise features \(\bm{F}_a \in \mathbb{R}^{(C+1) \times D} \) as
\begin{align} \label{eq:class_wise_attention}
\bm{A}_a(t, j) &= \frac{\exp(\tau \cdot \bm{S}_a(t, j))}{\sum_k \exp(\tau \cdot \bm{S}_a(k, j))}, \\
\bm{F}_a(j) &= \sum\nolimits_t \bm{A}_a(t, j)\bm{X}_e(t), \label{eq:class_wise_features}
\end{align}
where \(t\) denotes the \(t^{th}\) snippet, \(j\) represents the \(j^{th}\) category, and \(\tau\) is a temperature hyper-parameter controlling the smoothness of the softmax function.
It is obvious that the feature \(\bm{F}_a(j)\) should be identified as the foreground if the \(j^{th}\) action is performed in the video. Conversely, if the \(j^{th}\) action is absent in the video, it should be classified as the background. This observation promotes us to introduce a foreground classification process for the feature \(\bm{F}_a\). Specifically, given the foreground classifier \(\bm{W}_{f}\), we can obtain the class-wise foreground activation scores \(\bm{R}_a \in \mathbb{R}^{C+1}\) and the class-wise foreground confidences \(\bm{P}_a \in \mathbb{R}^{C+1}\) as
\begin{align} \label{eq:cw_vid_cls_scr}
\bm{R}_a(j) &= \delta \cdot \cos(\bm{F}_a(j), \bm{W}_{f}), \\
\bm{P}_a(j) &= \frac{\exp(\bm{R}_a(j))}{\sum_i \exp(\bm{R}_a(i))}. \label{eq:cw_vid_cls_pred}
\end{align}
A normalized cross-entropy loss \(\mathcal{L}_{cw}\) is computed as
\begin{equation} \label{eq:cw_loss}
\mathcal{L}_{cw} = -\mathbb{E}[\hat{\bm{y}}_a^T \log \bm{P}_a],
\end{equation}
where \(\hat{\bm{y}}= \bm{y} / \sum\nolimits_{i=1}^{C+1} y(i) \) is the normalized ground-truth vector, and \(\bm{y}(C+1)=0\). At this time, we actually transform the multi-label classification problem into multiple binary classification problems.
\vspace{-4mm}
\paragraph{Discussion:} Even if the CW branch is simple, it plays an important role in foreground-background separation. Specifically, we can transform Eq. \eqref{eq:cw_vid_cls_pred} into
\begin{equation} \label{eq:class_wise_softmax_expand}
\bm{P}_a(j) = \frac{\exp(\delta \cdot \cos_j)}{\exp(\delta \cdot \cos_j) + \sum_{i, i \neq j} \exp(\delta \cdot \cos_i)},
\end{equation}
where \(\cos_j\) is a simplification of \(\cos(\bm{F}_a(j), \bm{W}_{f})\). If the action category \(i\) (\(i\neq j\)) is absent in the video, it is expected that the feature \(\bm{F}_a(i)\) is a background feature. So there is one positive sample (if there is only one category in the video) from foreground and \(C\) negative samples from background. Therefore, Eq. \eqref{eq:class_wise_softmax_expand} is similar to the noise contrastive estimation (NCE) \cite{gutmann2010noise,oord2018representation,he2020momentum} process, minimizing Eq. \eqref{eq:cw_loss} actually maximizes the lower bound of mutual information (MI) between the foreground classifier \(\bm{W}_f\) and the feature \(\bm{F}_a(j)\). Besides, the background features are sampled from the same video of the feature \(\bm{F}_a(j)\), they can be viewed as hard negative samples, because action instances are usually surrounded by visually similar clips \cite{liu2019completeness}, which further guarantees the foreground-background separation. Therefore, the CW branch does not only introduce the action-to-foreground relation into our method but also enable the learning of robust and discriminative features.
However, the above analysis should be based on a meaningful foreground feature, but with the background class, there would be an ambiguity between the feature \(\bm{W}_f\) of the foreground classifier and the background feature \(\bm{W}_a(C+1)\), leading to inferior performance as shown in our experiments. Therefore, it is essential to enhance the foreground meaning of the \(\bm{W}_f\).
Besides, the CW branch only considers the unilateral relation from actions to foreground, which is insufficient to build the foreground-action consistency.

\subsection{Class-agnostic Attention Branch} \label{sec:ca_branch}
To complement the loss of the relation from foreground to actions, we adopt a class-agnostic attention branch (CA branch), which also enables to learn a semantically meaningful foreground classifier \(\bm{W}_{f}\), playing a complementary role with the CW branch. We first calculate the frame-wise foreground activation scores \(\bm{S}_f \in \mathbb{R}^{T}\) to obtain the foreground attention scores \(\bm{A}_f \in \mathbb{R}^{T}\) as:
\begin{align} \label{eq:foreground_score}
\bm{S}_f(t) &= \delta \cdot \cos(\bm{X}_e(t), \bm{W}_{f}), \\
\bm{A}_f(t) &= \frac{\exp(\tau \cdot \bm{S}_f(t))}{\sum_k \exp(\tau \cdot \bm{S}_f(k))} . \label{eq:foreground_attention}
\end{align}
Likewise, we can obtain a video-specific foreground feature
\(\bm{F}_f \in \mathbb{R}^{D}\) through a feature aggregation process
\begin{equation}
\bm{F}_f = \sum\nolimits_t \bm{A}_f(t)\bm{X}_e(t).
\end{equation}
Then we calculate the cosine similarity between the feature \(\bm{F}_f\) and the action classifier \(\bm{W}_a\) to obtain the video-level class confidence scores \(\bm{P}_f \in \mathbb{R}^{C+1}\)
\begin{equation} \label{eq:ca_vid_cls_pred}
\bm{P}_f(j) = \frac{\exp(\delta \cdot \cos(\bm{F}_f, \bm{W}_a(j)))}{\sum_i \exp(\delta \cdot \cos(\bm{F}_f, \bm{W}_a(i)))}.
\end{equation}
A normalized cross-entropy loss \(\mathcal{L}_{ca}\) is computed as
\begin{equation} \label{eq:ca_loss}
\mathcal{L}_{ca} = -\mathbb{E}[\hat{\bm{y}}^T \log \bm{P}_f],
\end{equation}
where \(\hat{\bm{y}} \) is the same as the CW branch, \ie, \(\bm{y}(C+1)=0\). In this way, the CA branch is exactly a symmetric pipeline with the CW branch, which is also consistent with the inverse relations they introduce.

\subsection{Multiple Instance Learning Branch} \label{sec:mil_branch}
In addition to the class-agnostic attention pipeline, the multiple instance learning (MIL) pipeline is also a good complement to the CW branch. First, the MIL pipeline also considers the unilateral relation from foreground to actions. Second, the temporal top-\(k\) average pooling of the MIL is actually a class-wise hard attention operation, which could help to better learn the class-wise attention score in the CW branch. Besides, the MIL is more concerned with whether a class occurs in the whole video, while the class-agnostic attention focuses more on the local (because the aggregation is linear for each frame), the two pipelines provide two different perspectives to classify a video and would be complementary in a way.
To better introduce the MIL pipeline into our framework, we change the temporal top-\(k\) average pooling into a class-wise soft attention operation, \ie, we share the class-wise attention scores \(\bm{A}_a \) (Eq. \eqref{eq:class_wise_attention}) and then aggregate the frame-wise class activation scores \(\bm{S}_a\) (Eq. \eqref{eq:class_similaritiy}) into the video-level class activation scores \(\bm{R}_m \in \mathbb{R}^{C+1}\) as
\begin{equation} \label{eq:mil_vid_cls_scr}
\bm{R}_m(j) = \sum\nolimits_t \bm{A}_a(t, j)\bm{S}_a(t, j).
\end{equation}
Similar to the CW branch and the CA branch, we can obtain its corresponding prediction \(\bm{P}_m \in \mathbb{R}^{C+1}\) and a normalized cross-entropy loss \(\mathcal{L}_{mil}\). Note that, since the background snippets exist in all videos, the ground truth \(\bm{y}\) of the MIL branch should have the background, \ie, \(\bm{y}(C+1)=1\).

\subsection{Hybrid Attention} \label{sec:hybrid_attention}
As we can see, the attention mechanism plays an important role in our framework. However, even though we strive to build the foreground-action consistency, we find that the attention scores still cannot well cover the ground truth. Just as stated in \cite{liu2019completeness}, attention scores are prone to focus on discriminative foreground snippets and visually similar background snippets. In order to address this issue and simultaneously maintain the attention-based structure, we propose a hybrid attention strategy. The motivation comes from a similar observation with \cite{nguyen2018weakly}, \ie, an action can be recognized by identifying a set of key frames. If we can make the attention scores focus on key frames, false positives are expected to be largely reduced. To achieve this, we utilize a simple but effective way that uses a large temperature hyper-parameter \(\tau\) for attention generation (\eg, Eq. \eqref{eq:class_wise_attention} and Eq. \eqref{eq:foreground_attention}).
In this way, the attention scores would be concentrated around snippets of high confidences.
\begin{figure}[!t]
\centering
\includegraphics[width=1\columnwidth]{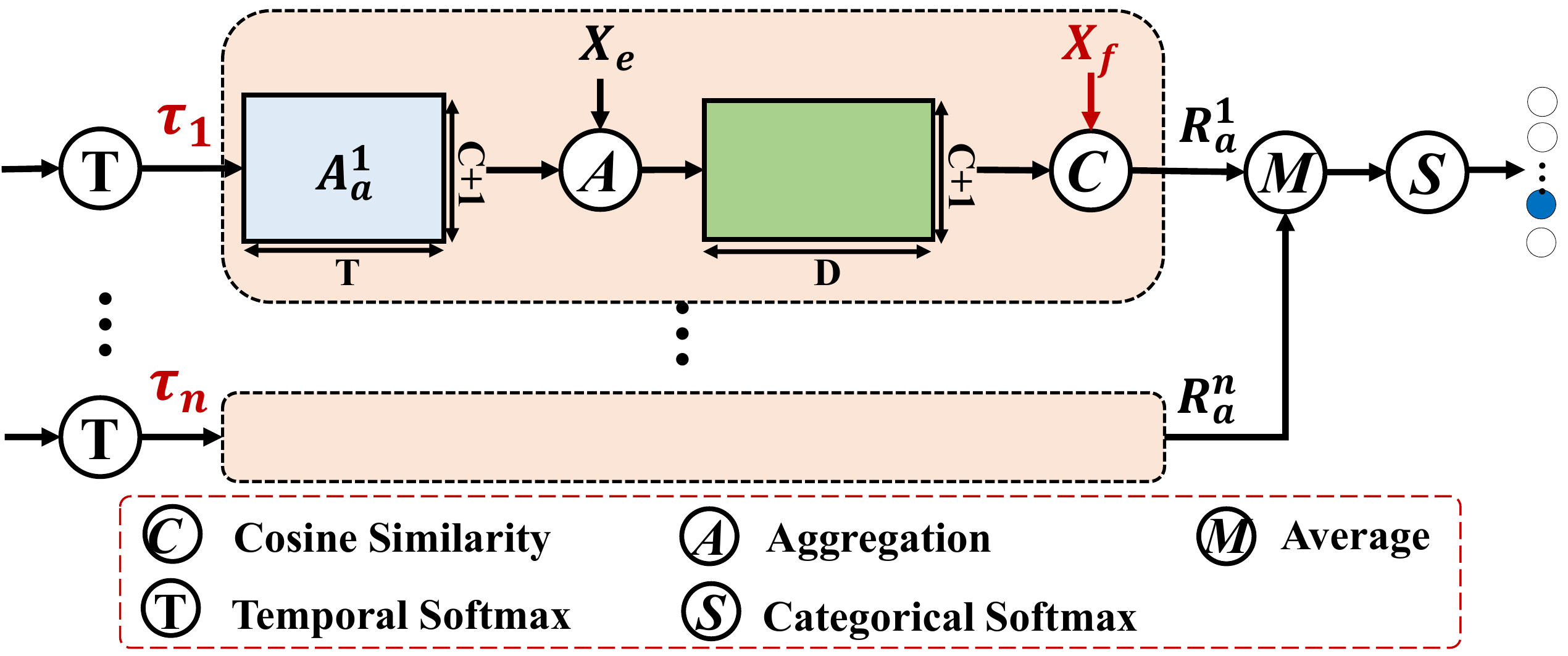}
\caption{Illustration of the hybrid attention strategy of the CW branch. We use \(N\) different \(\tau\) to calculate \(N\) class-wise attention scores \(\{\bm{A}_a^i\}_{i=1}^N\), the video-level foreground activation scores \(\{\bm{R}_a^i\}_{i=1}^N\) are averaged to obtain the final foreground activation score. The other two branches use this strategy in the same way.}
\label{fig:hybrid_attention}
\vspace{-4mm}
\end{figure}

Nevertheless, considering only key frames is insufficient, false negatives can also increase. Fortunately, because the key frames are modeled by the new attention scores (\(\tau>1.0\)), the original attention scores (\(\tau=1.0\)) have to accommodate some less-discriminative action snippets, which provides a fallback mechanism for our method. Therefore, using a hybrid attention strategy would be a reasonable way to improve the performance. As show in Fig. \ref{fig:hybrid_attention}, for the CW branch, we first use \(N\) different \(\tau\) to calculate \(N\) class-wise attention scores \(\{\bm{A}_a^i\}_{i=1}^N\). With each class-wise attention score \(\bm{A}_a^i\), we can obtain the video-level foreground activation score \(\bm{R}_a^i\) (Eq. \eqref{eq:cw_vid_cls_scr}). Finally, we average \(\{\bm{R}_a^i\}_{i=1}^N\) and use a softmax operation (Eq. \eqref{eq:cw_vid_cls_pred}) to obtain the probability confidence scores. The other two branches use the hybrid attention strategy in the same way.

\subsection{Training Objectives} \label{sec:training_objective}
Our model is jointly optimized with three video-level classification losses. The overall loss function is as follows:
\begin{equation} \label{eq:total_loss}
\mathcal{L}_{total} = \lambda_{cw} \mathcal{L}_{cw} + \lambda_{ca} \mathcal{L}_{ca} + \lambda_{mil} \mathcal{L}_{mil}
\end{equation}
where \(\lambda_{cw}\), \(\lambda_{ca}\) and \(\lambda_{mil}\) are balancing hyper-parameters. Our method can also work without the background class. At this time, we use the original ground truth \(\bm{y} \in \mathbb{R}^{C}\).
\begin{table*}[!t]
\begin{center}
\caption{Detection performance comparisons over the THUMOS14 dataset. The column AVG indicates the average mAP at IoU thresholds 0.1:0.1:0.7. UNT and I3D represent UntrimmedNet features and I3D features, respectively. $\dag$ means the method utilizes additional weak supervisions. \emph{FAC-Net w/o BG} indicates that our method do not use a background class.} \label{table:THUMOS14}
\begin{tabular}{c|c|l|ccccccc|c}
\hline
\hline
\multirow{2}{*}{Supervision} & \multirow{2}{*}{Year} & \multirow{2}{*}{Method} & \multicolumn{8}{c}{mAP @ IoU (\%)} \\
& & & 0.1 & 0.2 & 0.3 & 0.4 & 0.5 & 0.6 & 0.7 & AVG (0.1:0.7)  \\
\hline
\hline
\multirow{7}{*}{Full}
& 2017 & R-C3D \cite{xu2017r} & 54.5 & 51.5 & 44.8 & 35.6 & 28.9 & - & - & - \\
& 2018 & TAL-Net \cite{chao2018rethinking} & 59.8 & 57.1 & 53.2 & 48.5 & 42.8 & 33.8 & 20.8 & 45.1 \\
& 2018 & BSN \cite{lin2018bsn} & - & - & 53.5 & 45.0 & 36.9 & 28.4 & 20.0 & - \\
& 2019 & GTAN \cite{long2019gaussian} & 69.1 & 63.7 & 57.8 & 47.2 & 38.8 & - & - & - \\
\hline
\hline
\multirow{2}{*}{Weak $\dag$}
& 2018 & STAR (I3D) \cite{xu2018segregated} & 68.8 & 60.0 & 48.7 & 34.7 & 23.0 & - & - & - \\
& 2019 & 3C-Net (I3D) \cite{narayan20193c}  & 59.1 & 53.5 & 44.2 & 34.1 & 26.6 & - & 8.1 & - \\
\hline
\hline
\multirow{21}{*}{Weak}
& 2017 & UntrimmedNet \cite{wang2017untrimmednets} & 44.4 & 37.7 & 28.2 & 21.1 & 13.7 & - & - & - \\
& 2018 & STPN (I3D) \cite{nguyen2018weakly} & 52.0 & 44.7 & 35.5 & 25.8 & 16.9 & 9.9 & 4.3 & 27.0 \\
& 2018 & AutoLoc (UNT) \cite{shou2018autoloc} & - & - & 35.8 & 29.0 & 21.2 & 13.4 & 5.8 & - \\
& 2018 & W-TALC (I3D) \cite{paul2018w} & 55.2 & 49.6 & 40.1 & 31.1 & 22.8 & - & 7.6 & - \\
& 2019 & MAAN (I3D) \cite{yuan2019marginalized} & 59.8 & 50.8 & 41.1 & 30.6 & 20.3 & 12.0 & 6.9 &31.6 \\
& 2019 & CMCS (I3D) \cite{liu2019completeness} & 57.4 & 50.8 & 41.2 & 32.1 & 23.1 & 15.0 & 7.0 & 32.4 \\
& 2019 & BM (I3D) \cite{nguyen2019weakly} & 60.4 & 56.0 & 46.6 & 37.5 & 26.8 & 17.6 & 9.0 & 36.3 \\
& 2020 & BaS-Net (I3D) \cite{lee2019background} & 58.2 & 52.3 & 44.6 & 36.0 & 27.0 & 18.6 & 10.4 & 35.3 \\
& 2020 & RPN (I3D) \cite{huang2020relational} & 62.3 & 57.0 & 48.2 & 37.2 & 27.9 & 16.7 & 8.1 & 36.8 \\
& 2020 & TSCN (I3D) \cite{zhai2020two} & 63.4 & 57.6 & 47.8 & 37.7 & 28.7 & 19.4 & 10.2 & 37.8 \\
& 2020 & EM-MIL (I3D) \cite{luo2020weakly} & 59.1 & 52.7 & 45.5 & 36.8 & 30.5 & 22.7 & \textbf{16.4} & 37.7 \\
& 2020 & A2CL-PT (I3D) \cite{min2020adversarial} & 61.2 & 56.1 & 48.1 & 39.0 & 30.1 & 19.2 & 10.6 & 37.8 \\
& 2021 & HAM-Net (I3D) \cite{islam2021hybrid} & 65.4 & 59.0 & 50.3 & 41.1 & 31.0 & 20.7 & 11.1 & 39.8 \\
& 2021 & UM (I3D) \cite{lee2020background} & 67.5 & 61.2 & 52.3 & 43.4 & \textbf{33.7} & \textbf{22.9} & 12.1 & 41.9 \\
& - & FAC-Net w/o BG (I3D) & 63.8 & 57.5 & 48.1 & 40.5 & 31.3 & 20.0 & 10.2 & 38.8 \\
& - & FAC-Net (I3D) & \textbf{67.6} & \textbf{62.1} & \textbf{52.6} & \textbf{44.3} & 33.4 & 22.5 & 12.7 & \textbf{42.2} \\
\hline
\hline
\end{tabular}
\end{center}
\vspace{-6mm}
\end{table*}

\section{Experiments}
\subsection{Datasets}
We evaluate our method on two action localization datasets THUMOS14 \cite{THUMOS14} and ActivityNet1.3 \cite{caba2015activitynet}. Note that, we only use video-level category labels for training.

\vspace{-5mm}
\paragraph{THUMOS14.}
We use the subset from THUMOS14 that offers frame-wise annotations for 20 classes. We train the model on 200 untrimmed videos in its validation set and evaluate it on 212 untrimmed videos from the test set.

\vspace{-5mm}
\paragraph{ActivityNet1.3.}
This dataset covers 200 complex daily activities and provides 10,024 videos for training, 4,926 for validation and 5,044 for testing.
We use the training set to train our model and the validation set to evaluate our model.

\subsection{Implementation Details}
\vspace{-1mm}
\paragraph{Model details.} We use I3D \cite{carreira2017quo} for feature extraction. The network for extracting task-oriented features contains two layers, the output channels are 1024 and 1024, respectively. The scale factor \(\delta\) of cosine similarity is set as 5.0.
We utilize the hybrid attention strategy in a three-head way, with temperature hyper-parameters of 1.0, 2.0 and 5.0. We use ReLU as the activation function in our model, and dropout layers are utilized before all activation functions.

\vspace{-5mm}
\paragraph{Training details.}
Our method is implemented with PyTorch \cite{paszke2017automatic}.
During training, we loop through each video in the mini-batch and accumulate gradients to deal with variable video lengths.
We use Adam \cite{kingma2014adam} to optimize our model, the training procedure stops at 100 epochs with the learning rate 0.0001. The balancing hyper-parameters \(\lambda_{cw}\), \(\lambda_{ca}\) and \(\lambda_{mil}\) are 1.0, 0.1 and 0.1, respectively.

\vspace{-5mm}
\paragraph{Testing details.}
We take the whole sequence of a video as input for testing. When localizing action instances, the class activation sequence is upsampled to the original frame rate. We reject the category whose class probability \(\bm{P}_f(j)\) (Eq. \eqref{eq:ca_vid_cls_pred}) is lower than 0.1.
Following \cite{lee2019background}, we use a set of thresholds to obtain the predicted action instances, then we perform non-maximum suppression to remove overlapping segments among rgb stream and optical-flow stream.

\subsection{Comparison with The State-of-the-art}
As shown in Tab. \ref{table:THUMOS14}, on THUMOS14, even though we do not adopt a background class (\ie, \emph{FAC-Net w/o BG}), our method still outperforms the existing background modeling approaches \cite{nguyen2019weakly,lee2019background}, indicating the effectiveness of building foreground-action consistency. Besides, with the background class, our method obtains a new state-of-the-art performance, achieving gains in terms of mAPs at most IoU threholds and average mAP. Notably, our method outperforms some fully-supervised methods at IoU 0.1 and 0.2, manifesting the potential of weakly-supervised method.

Tab. \ref{table:activity1.3} demonstrates the results on the ActivityNet1.3 dataset. As we can see, despite the simple architecture, our method obtains comparable performance with state-of-the-art approaches, and surpasses the fully-supervised methods R-C3D \cite{xu2017r} and TAL-Net \cite{chao2018rethinking} by large margins of 11.3\% and 3.8\% in terms of average mAP, respectively.

\subsection{Ablation Studies}
We conduct a set of ablation studies on THUMOS14.
\begin{table}[!t]
\begin{center}
\caption{Results on ActivityNet1.3 validation set. AVG indicates the average mAP at IoU thresholds 0.5:0.05:0.95.}
\label{table:activity1.3}
\begin{tabular}{l|cccc}
\hline
\hline
\multirow{2}{*}{Method} & \multicolumn{4}{c}{mAP @ IoU} \\
\cline{2-5}
& 0.5 & 0.75 & 0.95 & AVG \\
\hline
\hline
R-C3D \cite{xu2017r} & 26.8 & - & - & 12.7 \\
TAL-Net  \cite{chao2018rethinking} & 38.2 & 18.3 & 1.3 & 20.2\\
\hline
CMCS (I3D) \cite{liu2019completeness} & 34.0 & 20.9 & 5.7 & 21.2\\
MAAN (I3D) \cite{yuan2019marginalized} & 33.7 & 21.9 & 5.5 & -\\
BaS-Net (I3D) \cite{lee2019background} & 34.5 & 22.5 & 4.9 & 22.2\\
A2CL-PT (I3D) \cite{min2020adversarial} & 36.8 & 22.0 & 5.2 & 22.5\\
TSCN (I3D) \cite{zhai2020two} & 35.3 & 21.4 & 5.3 & 21.7\\
UM (I3D) \cite{lee2020background} & 37.0 & 23.9 & 5.7 & 23.7\\
ACM-BANet (I3D) \cite{moniruzzaman2020action} & \textbf{37.6} & {24.7} & \textbf{6.5} & \textbf{24.4}\\
FAC-Net (I3D) & \textbf{37.6} & 24.2 & 6.0 & 24.0 \\
\hline
\hline
\end{tabular}
\vspace{-7mm}
\end{center}
\end{table}

\vspace{-5mm}
\paragraph{Branch analysis.} To figure out the contribution of each branch, we should consider two questions: What is the performance of the individual branch? What is the relationship between different branches? In Tab. \ref{table:ablation}, even if each of the three branches only considers unilateral relation between foreground and actions, the CW branch obtains better performance (31.2\%), indicating that the relation from actions to foreground is more important, which can also enforce the foreground-background separation. Moreover, combining any two branches can consistently improve the performance, demonstrating the complementary relations among the three branches. Especially, the CW branch can significantly boost the performance of the CA branch and the MIL branch by 5.4\% and 4.8\% on average mAP, respectively.
\begin{figure}[!t]
\centering
\includegraphics[width=1\columnwidth]{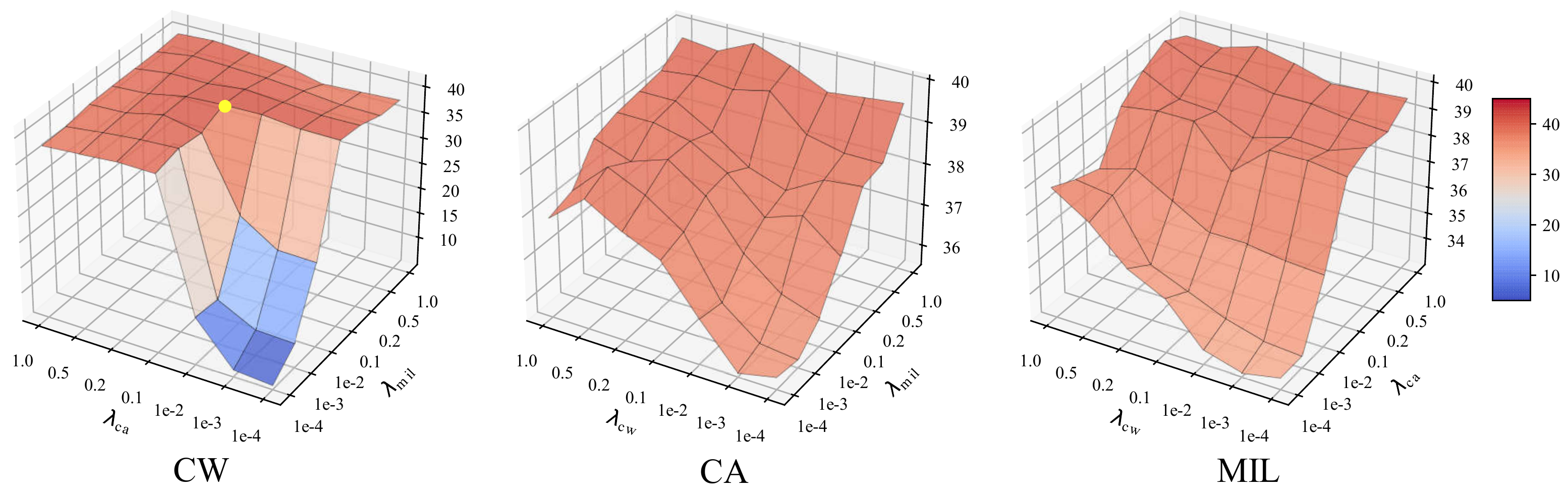}
\caption{Ablation study on the balancing hyper-parameters \(\lambda_{cw}\), \(\lambda_{ca}\) and \(\lambda_{mil}\). We use a background class in these experiments.}
\label{fig:hyperparameters}
\vspace{-4mm}
\end{figure}

In Fig. \ref{fig:hyperparameters}, we further explore the relations among branches by controlling the balancing hyper-parameters \(\lambda_{cw}\), \(\lambda_{ca}\) and \(\lambda_{mil}\). For example, when we evaluate the CW branch, we fix the \(\lambda_{cw}\) as 1.0, and adjust the balancing hyper-parameters of other two branches. Note that, we use a background class in these experiments. We can find that, when \(\lambda_{ca}\) and \(\lambda_{mil}\) are small, the performance drops to a very low level (about 6\%). The reason is that the foreground classifier \(\bm{W}_f\) along with the background class arouse an ambiguity of foreground and background, leading to inferior performance.
When \(\lambda_{ca}\) and \(\lambda_{mil}\) are large, the performance is insensitive to the \(\lambda_{cw}\), and remains at a promising level (above 40.0\%).
Besides, taking the CW branch as the main branch obtains the optimal result, even if it is used as an auxiliary branch, it also improves the performance.

\begin{figure}[!t]
\centering
\includegraphics[width=1\linewidth]{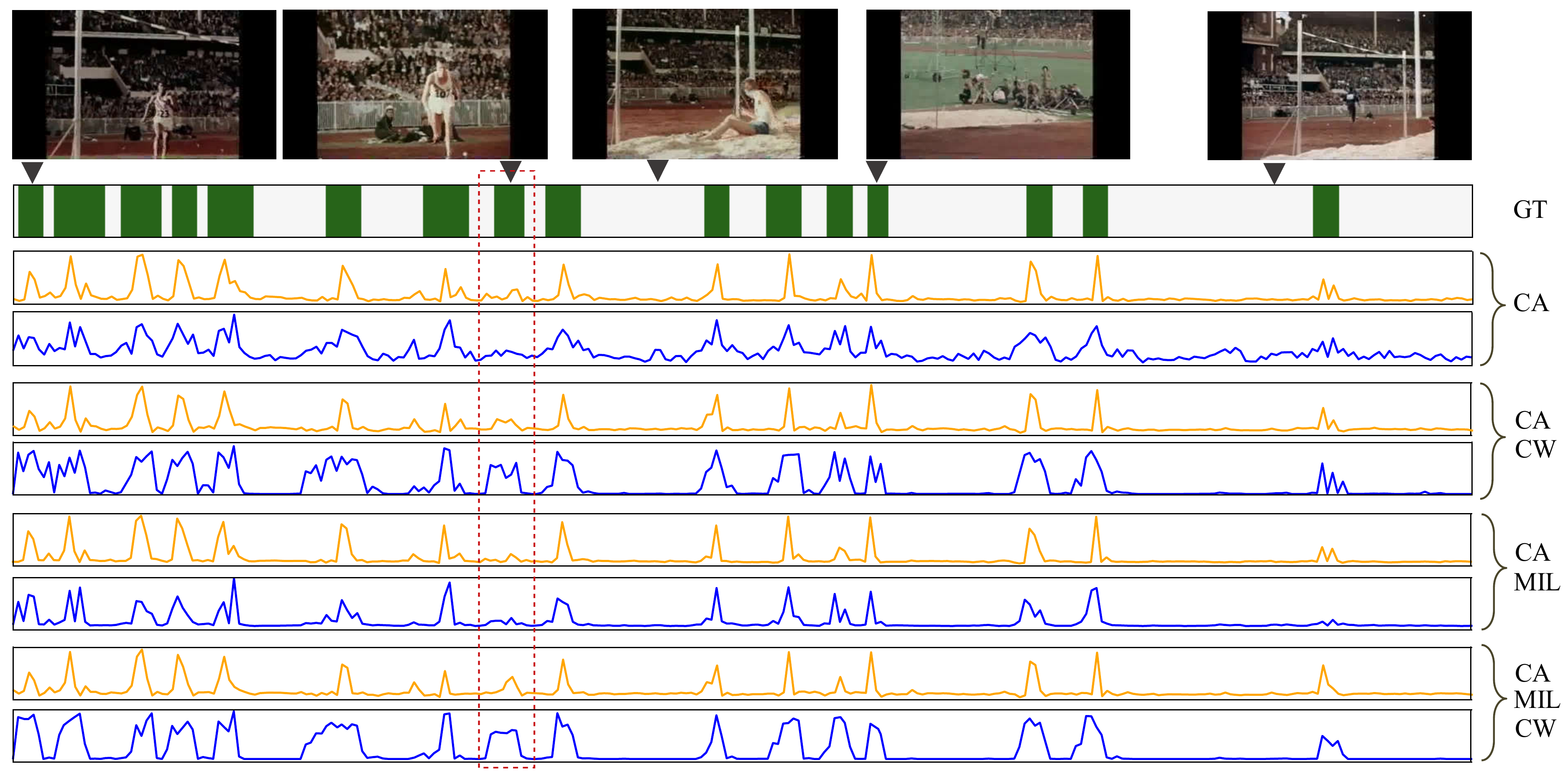}
\caption{Visualization of foreground activation scores and action activation scores under different model setting. We show an example of ``\emph{Pole Vault}'' in THUMOS14. For fair comparison, we do not use a background class.}
\label{fig:score_visualization}
\vspace{-2mm}
\end{figure}

To attain an intuitive insight into the three branches, in Fig. \ref{fig:score_visualization}, we visualize the foreground activation scores (Eq. \eqref{eq:foreground_score}) and action activation scores of ground truth under different combinations of branches. After adding the CW branch, foreground activation scores can better cover the ground truth, leading to more accurate detection results.
\begin{table}[!t]
\begin{center}
\caption{Ablation studies on the THUMOS14 dataset. The column AVG indicates the average mAP at IoU thresholds 0.1:0.7.}
\label{table:ablation}
\vspace{-3mm}
\resizebox{1\columnwidth}{!}{
\begin{tabular}{ccccc|c}
\hline
\hline
\multicolumn{5}{c|}{\textbf{Method}} & \multirow{2}{*}[-0.7em]{\makecell{AVG\\(0.1:0.7)}} \\
\cline{1-5}
\rule{0pt}{12pt} \makecell{CW\\Branch} & \makecell{CA\\Branch} & \makecell{MIL\\Branch} & \makecell{Hybrid\\Attention} & \makecell{Background\\Class}\\
\hline
\hline
\CheckmarkBold &  &  &  &  & 31.2 \\
& \CheckmarkBold  &  &  &  & 30.5 \\
&  & \CheckmarkBold  &  &  & 29.6 \\
\CheckmarkBold &  &  & \CheckmarkBold &  & 34.6 \\
& \CheckmarkBold  &  & \CheckmarkBold  &  & 32.5 \\
&  & \CheckmarkBold  & \CheckmarkBold  &  & 32.5 \\
\hline
\CheckmarkBold & \CheckmarkBold &  &  &  & 35.9 \\
\CheckmarkBold & \CheckmarkBold &  & \CheckmarkBold &  & 36.5 \\
\CheckmarkBold &  & \CheckmarkBold &  &  & 34.4 \\
\CheckmarkBold &  & \CheckmarkBold & \CheckmarkBold &  & 35.7 \\
& \CheckmarkBold & \CheckmarkBold &  &  & 35.7 \\
& \CheckmarkBold & \CheckmarkBold & \CheckmarkBold &  & 36.3 \\
\CheckmarkBold & \CheckmarkBold & \CheckmarkBold &  &  & 37.5 \\
\CheckmarkBold & \CheckmarkBold & \CheckmarkBold & \CheckmarkBold &  & 38.8 \\
\hline
\CheckmarkBold &  & \CheckmarkBold &  & \CheckmarkBold & 38.3 \\
\CheckmarkBold &  & \CheckmarkBold & \CheckmarkBold & \CheckmarkBold & 39.8 \\
& \CheckmarkBold & \CheckmarkBold &  & \CheckmarkBold & 37.6 \\
& \CheckmarkBold & \CheckmarkBold & \CheckmarkBold & \CheckmarkBold & 38.4 \\
\CheckmarkBold & \CheckmarkBold & \CheckmarkBold &  & \CheckmarkBold & 40.8 \\
\CheckmarkBold & \CheckmarkBold & \CheckmarkBold & \CheckmarkBold & \CheckmarkBold & \textbf{42.2} \\
\hline
\hline
\end{tabular}}
\vspace{-5mm}
\end{center}
\end{table}

\vspace{-5mm}
\paragraph{Effect of hybrid attention.}
From Tab. \ref{table:ablation}, we can see that the hybrid attention can consistently improve the performance, especially for the single branch. Besides, Tab. \ref{table:hybrid_attention_study} shows the ablation studies on the number of attentions and temperature hyper-parameters \(\tau\). We can find that the number of attentions is not the more the better, too many attentions would degenerate the performance. Likewise, a large \(\tau\), \eg, 10.0, also works against the model, making it to focus too much on discriminative snippets.
\begin{figure}[!t]
\centering
\includegraphics[width=1\linewidth]{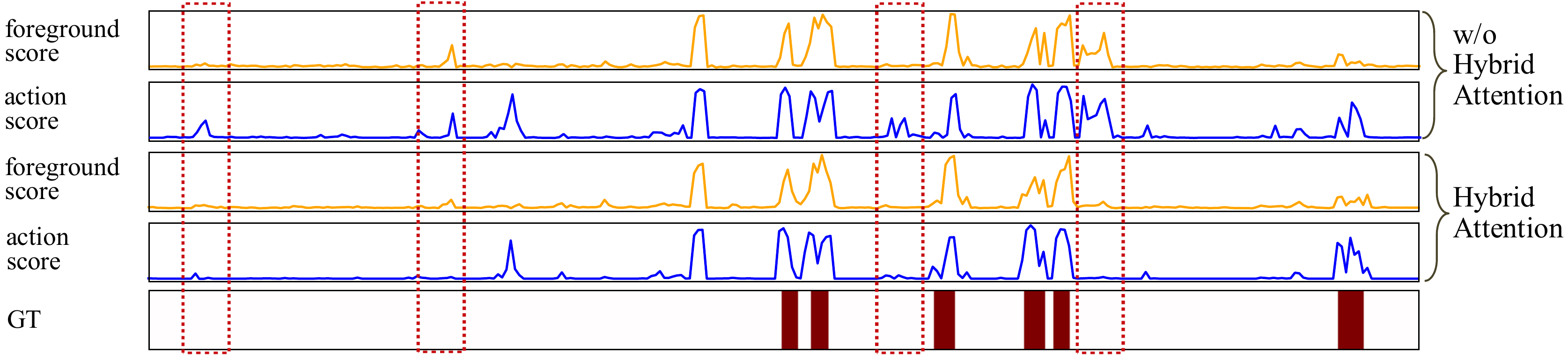}
\caption{Visualization of foreground activation scores and action activation scores without and with the hybrid attention strategy. We show an example of ``\emph{Cliff Diving}'' in THUMOS14.}
\label{fig:hybrid_attention_visualization}
\vspace{-3mm}
\end{figure}
\begin{figure*}[!t]
\centering
\includegraphics[width=1\linewidth]{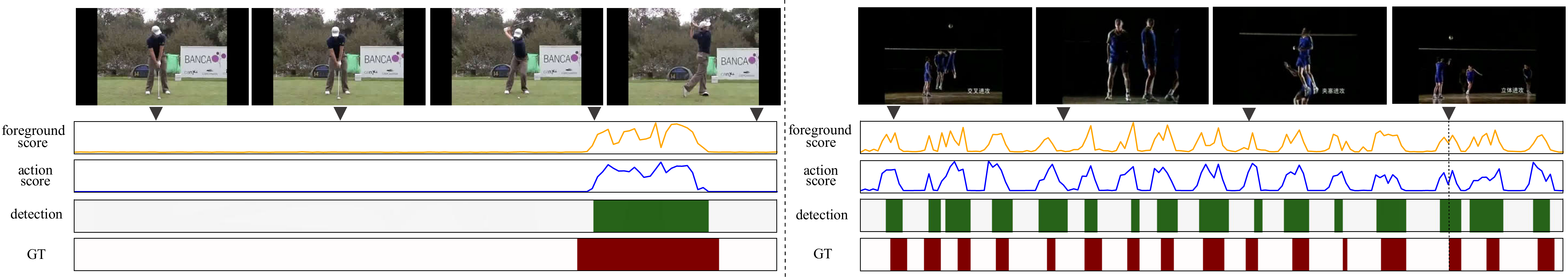}
\caption{Qualitative results on THUMOS14 \cite{THUMOS14}. We show: 1) foreground activation scores, 2)  activation scores of ground truth action, 3) the detected action instances and 4) ground truth.
\textbf{Left:} An example of \emph{Golf Swing}. \textbf{Right:} An example of \emph{Volleyball Spiking}.}
\label{fig:detection_visualization}
\vspace{-4mm}
\end{figure*}
\begin{figure}[!t]
\centering
\includegraphics[width=0.95\linewidth]{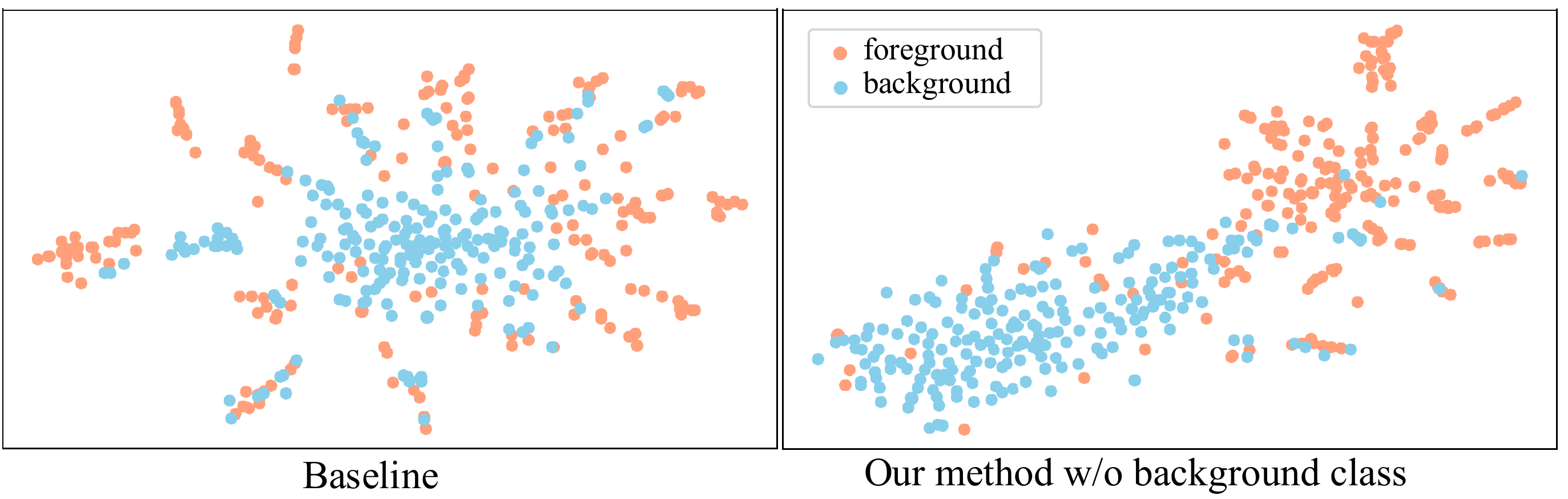}
\caption{Visualization of foreground-background separation of embedding features of (a) only the CA branch and (b) our method via t-SNE \cite{maaten2008visualizing} on the THUMOS14 test set. Note that, for a fair comparison, we do not use a background class in both models.}
\label{fig:feature_visualization}
\vspace{-2mm}
\end{figure}

In Fig. \ref{fig:hybrid_attention_visualization}, we also visualize the foreground activation scores and action activation scores under different hybrid attention settings. It is obvious that the introduction of the hybrid attention enables to obtain more accurate foreground prediction and effectively reduce false positives.

\vspace{-4mm}
\paragraph{Complementary role of the CW branch.} In light of the observation that existing methods ignore the action-to-foreground relation, intuitively, the CW branch may play a complementary role to existing methods. In Tab. \ref{table:complementary_study}, we plug the CW branch into four weakly-supervised methods. We can find that the CW branch can significantly boost the performance of two classic methods STPN \cite{nguyen2018weakly} and W-TALC \cite{paul2018w} by 1.7\% and 1.9\%, respectively. Even though the BM \cite{nguyen2019weakly} explicitly models the background, the CW branch can further improve its performance. Besides, the performance of the recent method UM \cite{lee2020background} can be also improved.
\begin{table}[!t]
\begin{center}
\caption{Evaluation of the hybrid attention on THUMOS14. AVG indicates the average mAP at IoU thresholds 0.1:0.1:0.7. Each line shows a hybrid attention setting, \eg, ``(2) $\sim$ 1.0, 2.0'' represents we use two attention scores with temperature hyper-parameter of 1.0 and 2.0, respectively.}
\vspace{1mm}
\label{table:hybrid_attention_study}
\begin{tabular}{l|cccc}
\hline
\hline
\multirow{2}{*}[-0.1em]{Hybrid Attention} & \multicolumn{4}{c}{AVG} \\
\cline{2-5}
& CW & CA & MIL & Full\\
\hline
\hline
(1) $\sim$ 1.0 & 31.2 & 30.5 & 29.6 & 40.8 \\
(2) $\sim$ 1.0, 2.0 & 32.0 & 32.1 & 31.7 & 41.6 \\
(2) $\sim$ 1.0, 3.0 & 33.6 & 32.3 & 31.6 & 41.8 \\
(2) $\sim$ 1.0, 5.0 & \textbf{34.7} & 31.7 & 32.0 & 41.9 \\
(2) $\sim$ 1.0, 10.0 & 32.9 & 31.2 & 26.4 & 33.8 \\
(3) $\sim$ 1.0, 2.0, 3.0 & 34.4 & 32.2 & 31.4 & 41.9 \\
(3) $\sim$ 1.0, 2.0, 5.0 & 34.6 & \textbf{32.5} & \textbf{32.5} & \textbf{42.2} \\
(3) $\sim$ 1.0, 5.0, 10.0 & 33.4 & 32.0 & 26.7 & 33.2 \\
(4) $\sim$ 1.0, 2.0, 3.0, 5.0 & 33.6 & 31.6 & 29.1 & 42.0 \\
\hline
\hline
\end{tabular}
\vspace{-3mm}
\end{center}
\end{table}

\subsection{Qualitative Results}
We visualize some examples of detected action instances in Fig. \ref{fig:detection_visualization}. In the first example of \emph{Golf Swing}, our method pinpoints the only one action instance. In the second example of \emph{Volleyball Spiking}. Even though this action is frequently performed in the video, our method successful detects all the action instances, which shows the ability to handle dense action occurrence. As we can see, our method significantly suppresses the responses of background. Besides, the foreground scores and action scores are consistent and well cover the ground truth.
Fig. \ref{fig:feature_visualization} shows the visualization of the features \(\bm{X}_e\) about their foreground-background separation. As we can see, our method can better separate foreground from background than the baseline model.
\begin{table}[!t]
\begin{center}
\caption{Evaluation of the complementary role of the CW branch. Note that, our method needs to learn task-oriented features, which is impossible in STPN, so we show the result of ``STPN + Embedding'', which represents STPN plus a feature embedding module.}
\vspace{-2mm}
\label{table:complementary_study}
\resizebox{1\columnwidth}{!}{
\begin{tabular}{c|l|ccc}
\hline
\hline
\multicolumn{2}{c|}{\multirow{2}{*}{Method}} & \multicolumn{3}{c}{mAP @ IoU} \\
\cline{3-5}
\multicolumn{2}{c|}{} & 0.3 & 0.7 & AVG \\
\hline
\hline
\multirow{5}{*}{\makecell{Attention\\based}}
& STPN \cite{nguyen2018weakly} (reproduced) & 35.2 & 4.2 & 26.8 \\
& STPN + Embedding & 38.4 & 4.7 & 28.9 \\
& STPN + Embedding + CW  & 40.3 & 5.6 & 30.6 \\
& BM \cite{nguyen2019weakly} (reproduced) & 46.5 & 9.1 & 36.0\\
& BM + CW  & 47.9 & 10.0 & 37.6\\
\hline
\multirow{4}{*}{\makecell{MIL\\based}}
& W-TALC \cite{paul2018w} (reproduced) & 40.4 & 7.2 & 31.6\\
& W-TALC + CW & 42.0 & 8.7 & 33.5\\
& UM \cite{lee2020background} (reproduced) & 51.0 & 10.9 & 40.4\\
& UM + CW & 51.6 & 11.1 & 40.8\\
\hline
\hline
\end{tabular}}
\vspace{-5mm}
\end{center}
\end{table}

\section{Conclusion}
We proposed a weakly-supervised action localization approach, named FAC-Net that consists of three branches.
Different from the existing methods that only considers unilateral relation from foreground to actions, our method takes bilateral relations between actions and foreground into consideration. The proposed class-wise foreground classification branch introduces the action-to-foreground relation to maximize
the foreground-background separation. Besides, the class-agnostic attention branch and the multiple instance learning branch are adopted to regularize the foreground-action consistency and learn a meaningful foreground feature.
According to our experiments, the class-wise foreground classification branch can play a complementary role to existing methods to improve their performance.

\section*{Acknowledgement}
This work is supported in part by Centre for Perceptual and Interactive Intelligence Limited, in part by the General Research Fund through the Research Grants Council of Hong Kong under Grants (Nos. 14204021, 14208417, 14207319, 14202217, 14203118, 14208619), in part by Research Impact Fund Grant No. R5001-18, in part by CUHK Strategic Fund.

{\small
\bibliographystyle{ieee_fullname}
\bibliography{egbib}
}

\end{document}